# A Case Study in Knowledge Discovery and Elicitation in an Intelligent Tutoring Application

Ann Nicholson* Tal Boneh† Tim Wilkin‡ Kaye Stacey§ Liz Sonenberg¶ Vicki Steinle‖

*School of Computer Sci. and Soft. Eng., Monash University, VIC 3800, Australia. annn@csse.monash.edu.au

† Department of Computer Science, The University of Melbourne, Parkville, VIC 3052, Australia. boneh@students.cs.mu.oz.au

‡As for A. Nicholson. taw@csse.monash.edu.au

§ Department of Science and Mathematics Education, The University of Melbourne, VIC 3010, Australia. k.stacey@unimelb.edu.au

¶ Department of Information Systems, The University of Melbourne, Parkville, VIC 3052, Australia. LizS@staff.dis.unimelb.edu.au

‖As for K. Stacey. v.steinle@edfac.unimelb.edu.au

## Abstract

Most successful Bayesian network (BN) applications to date have been built through knowledge elicitation from experts. This is difficult and time consuming, which has lead to recent interest in automated methods for learning BNs from data. We present a case study in the construction of a BN in an intelligent tutoring application, specifically decimal misconceptions. We describe the BN construction using expert elicitation and then investigate how certain existing automated knowledge discovery methods might support the BN knowledge engineering process.

## 1 INTRODUCTION

Bayesian networks have become a popular AI representation for reasoning under uncertainty, with successful applications in (medical) diagnosis, planning, monitoring, vision, information retrieval and intelligent tutoring [Conati et al., 1997, Mayo and Mitrovic, 2001, VanLehn and Niu, 2001]. Most successful applications to date have been built through knowledge elicitation from experts. In general, this is difficult and time consuming [Druzdzel and van der Gaag, 2001], with problems involving incomplete knowledge of the domain, common human difficulties in specifying and combining probabilities, and experts being unable to identify the causal direction of influences between variables. Hence there has been much interest in recent time in automated methods for constructing BNs from data (e.g., [Spirtes et al., 1993, Wallace and Korb, 1999, Heckerman and Geiger, 1995]). Most evaluation of these automated methods is done by taking an existing BN model, generating data from it that is given to the automated learner; the learned BN is compared to the original.

While there have been attempts to combine knowledge elicitation from experts and automated knowledge discovery methods (e.g. [Heckerman et al., 1994, Onisko et al., 2000]), there is as yet no established methodology [Kennett et al., 2001]. Appropriate evaluation, in particular, is an open question; most automated methods use some sort of statistical measure of how well the BN model fits the data whereas elicited models are assessed in part by how well their predictions on particular test scenarios meet expert expectations. When both data and expert knowledge about a domain is available, it is not simply a question of using the automated methods to validate the expert elicited BN, or using the expert to choose between learned networks or complete those not fully specified. Networks built using the different methods may be very different. The question then becomes how to resolve such differences such that the resultant BN model is acceptable to the expert/client and hence deployable.

In this paper, we present a case study in the construction of a BN model in the intelligent tutoring system (ITS) (Section 2). We describe the initial network construction using expert elicitation, together with a preliminary evaluation (Section 3). We then apply automated knowledge discovery methods to each main task in the construction process: (Section 4): (1) we apply a classification method to student test data; (2) we perform simple parameter learning based on fre-quency counts to the expert BN structures and (3) we apply an existing BN learning program. In each case we compare the performance of the resultant network with the expert elicited networks, providing an insight into how elicitation and knowledge discovery might be combined in the BN knowledge engineering process.

## 2 THE ITS DOMAIN

Decimal notation is widely used in our society. Our testing [Stacey and Steinle, 1999] of 5383 students has indicated that less 70% of Year 10 students (age about 15 years) understand the notation well enough to reliably judge the relative size of decimals. On the other hand, more than 30% of Grade 5 students (age about 10 years) have mastered this important concept. Expertise grows only very slowly throughout the intervening years under normal instruction in our schools, and so an intelligent tutoring approach to this important topic is of interest. Students' understanding of decimal notation has been mapped using a short test, the Decimal Comparison Test (DCT), where the student is asked to choose the larger number from each of 24 pairs of decimals [Stacey and Steinle, 1999]. The pairs of decimals are carefully chosen so that from the patterns of responses, students' (mis)understanding can be diagnosed as belonging to one of a number of categories. These categories have been identified manually, based on extensive research [Stacey and Steinle, 1999]. For most students, there is consistency in their responses to similar test items and some children display the same misconception over long periods of time.

About a dozen misconceptions have been identified using the DCT and interviews. Most are based on false analogies, which are sometimes embellished by isolated learned facts (such as a zero in the tenths column makes a number small). For example, many younger students think 0.4 is smaller than 0.35 because there are 4 parts (of unspecified size, for these students) in the first number and 35 parts (also of unspecified size) in the second. However, these students (LWH in Table 1) get many items right, e.g. 5.736 compared with 5.62, with the same erroneous thinking. Students in the SRN class (Table 1) choose 0.4 as greater than 0.35 but for the wrong reason, as they draw an analogy between fractions and decimals and use knowledge that 1/4 is greater than 1/35. See [Stacey and Steinle, 1999] for a detailed discussion of these responses and categories of students. Table 1 shows the rules the experts originally used to classify students based on their response to 6 types of DCT test items: H = High correctness (e.g. 4 or 5 out of 5), L = Low number correct (e.g. 0 or 1 out of 5), with '.' indicating that any performance level is observable for that item type by that student class other than the combinations seen above. We note that the fine misconception classifications have been "grouped" by the experts into a coarse classification – L (think longer decimals are larger numbers), S (shorter is larger), A (apparent expert) and UN (other). The LU, SU and AU fine classifications correspond to students who on their answers on Type 1 and 2 items behave like others in their coarse classification, however they don't behave like them on the other item types. These and the UNs may be students behaving consistently according to an unknown misconception, or students who are not following any consistent interpretation.

Table 1: Responses experts expect from students with different misconceptions.

| Expert Class | Item type | | | | | |
|---|---|---|---|---|---|---|
| | 1<br>0.4<br>0.35 | 2<br>5.736<br>5.62 | 3<br>4.7<br>4.08 | 4<br>0.452<br>0.45 | 5<br>0.4<br>0.3 | 6<br>0.42<br>0.35 |
| ATE | H | H | H | H | H | H |
| AMO | H | H | H | L | H | H |
| MIS | L | L | L | L | L | L |
| AU | H | H | . | . | . | . |
| LWH | L | H | L | H | H | H |
| LZE | L | H | H | H | H | H |
| LRV | L | H | L | H | H | L |
| LU | L | H | . | . | . | . |
| SDF | H | L | H | L | H | H |
| SRN | H | L | H | L | H | L |
| SU | H | L | . | . | . | . |
| UN | . | . | . | . | . | . |

We have developed an intelligent tutoring system based on computer games for this decimals domain[McIntosh et al., 2000]. The overall architecture of the system is shown in Figure 1. The computer game genre was chosen to provide children with an experience different from, but complementary to, normal classroom instruction and to appeal across the target age range (Grade 5 and above). Each game focuses on one aspect of decimal numeration, thinly disguised by a story line. It is possible for a student to be good at one game or the diagnostic test, but not good at another; emerging knowledge is often compartmentalised.[1]

---

[1]In the "Hidden Numbers" game students are confronted with two decimal numbers with digits hidden behind closed doors; the task is to find which number is the larger by opening as few doors as possible. The game "Flying Photographer" requires students to place a number on a number line, prompting students to think differently about decimal numbers. The "Number Between" game is also played on a number line, but particularly focuses on the density of the decimal numbers; students have to type in a number between a given pair. "Decimaliens" is a classic shooting game, designed to link various representations of the value of digits in a decimal number.

The simple expert rules classification described above makes quite arbitrary decisions about borderline cases. The use of a BN to model the uncertainty allows it to make more informed decisions in these cases. Using a BN also provides a framework for integrating student responses from the computer games with DCT information. The BN is initialised with a generic student model, with the options of individualising with classroom or online DCT results. The BN is used to update an ongoing assessment of the student's understanding, to predict which item types that student might be expected to get right or wrong, and, using sensitivity analysis, to identify which evidence would most improve the misconception diagnosis. The development of the BN is described below.

The system controller module uses the information provided by the BN, together with the student's previous responses, to select which item type to present to the student next, and to decide when to present help or change to a new game. This architecture allows flexibility in combining the teaching "sequencing tactic" (that is, whether easy items are presented before harder ones, harder first, or alternating easy/hard), coverage of all item types, and items which will most improve the diagnosis. More detailed descriptions of both the architecture and the item selection algorithm are given in [Stacey et al., 2001]. The ITS shown in Figure 1 (including the four computer games) has been fully implemented. The game interfaces are currently being assessed for usability on individual children, with deployment and assessment in the classroom environment to take place over the next year.

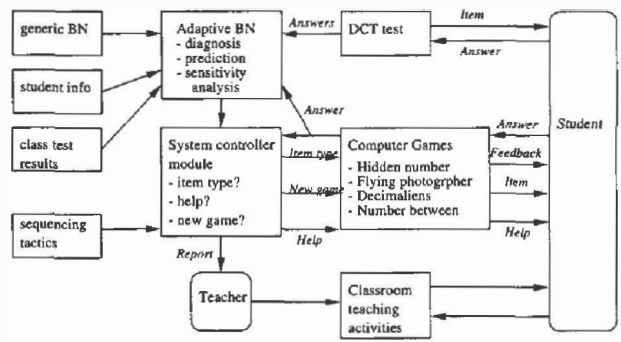

Figure 1: Intelligent Tutoring System Architecture

## 3 EXPERT ELICITATION

It is generally accepted that building a BN for a particular application domain involves three tasks [Druzdzel and van der Gaag, 2001]: (1) identification of the important variables, and their values; (2) identification and representation of the relationships between variables in the network structure; and (3) parameterisation of the network, that is determining the conditional probability tables associated with each network node. For our purposes we consider there to be an additional task (4) the evaluation of the network. While in theory these tasks can be performed sequentially, in practice the knowledge engineering process iterates over these task until the resultant network is considered "acceptable". In this section we describe the elicitation of the decimal misconception BN from the education domain experts.

### 3.1 BN VARIABLES

Student misconceptions are represented on two levels, by two variables. The coarseClass node can take the values L,S,A, and UN, whereas the fineClass node, incorporating all the misconception types identified by the experts, can take the 12 values shown in column 1 of Table 1. Note that the experts consider the classifications to be mutually exclusive. If that were not the case, then two variables would not be sufficient; rather we would require a Boolean variable for each of the classifications.

Each DCT item type is made a variable in the BN, representing student performance on test items of those types; student test answers are entered as evidence for these nodes. The following alternatives were considered for the possible values of the item type nodes.

1. Suppose the test contains N items of a given type. One possible set of values for the BN item type node is $\{0,1, ..., N\}$, representing the number of the items the student answered correctly. The number of items may vary for the different types, and for the particular test set given to the students, but it is not difficult to adapt the BN. Note that the more values for each node, the more complex the overall model; if N were large (e.g. > 20), this model may lead to complexity problems.

2. The item type node may be given the values {High, Medium, Low}, reflecting an aggregated assessment of the student's answers for that item type. For example, if 5 such items were presented, 0 or 1 correct would be considered low, 2 or 3 would be medium, while 4 or 5 would be High. For types with 4 items, medium encompasses only 2 correct, while for types with only 3 items, the medium value is omitted completely. This reflects the expert rules classification described above.

### 3.2 BN STRUCTURE

The experts considered the coarse classification to be a strictly deterministic combination of the fine classification, hence the coarseClass node was made a child of the fineClass node, For example, a student was considered an L if and only if it was one of an LWH, LZE, LRV or LU.

The type nodes are observation nodes, where entering evidence for a type node should update the posterior probability of a student having a particular misconception. This diagnostic reasoning is typically reflected in a BN structure where the class, or "cause" is the parent of the "effect" (i.e. evidence) node. Therefore an arc was added from the subclass node to each of the type nodes. No connections were added between any of the type nodes, reflecting the experts' intuition that a student's answers for different item types are independent, given the subclassification.

A part of the expert elicited BN structure implemented in the ITS is shown in Figure 2. This network fragment shows the coarseClass node (values L,S,A,UN), the detailed misconception fineClass node (12 values), the item type nodes used for the DCT, plus additional nodes for some games. These additional nodes are not described in this paper but are included to illustrate the complexity of the full network.[2] The bolded nodes are those in the restricted network used subsequently in this paper for evaluation and experimentation.

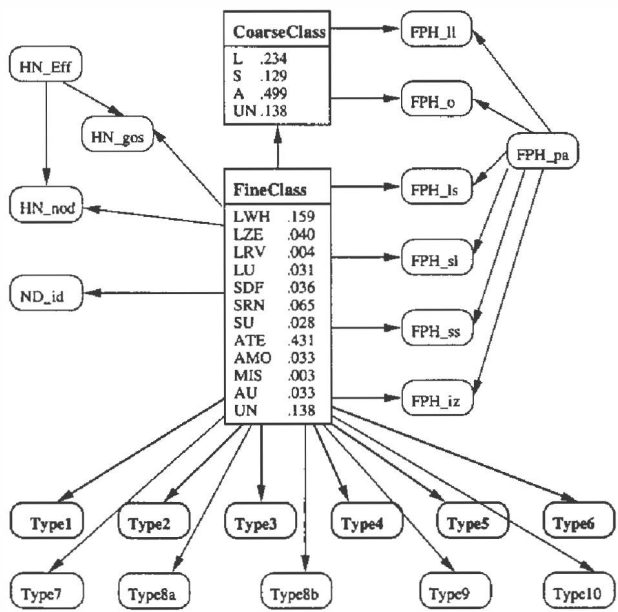

Figure 2: Fragment of the expert elicited BN currently implemented. Bold nodes are those discussed here.

### 3.3 BN PARAMETERS

The education experts had collected data that consisted of the test results and the expert rule classification on a 24 item DCT for over two thousand five hundred students from Grades 5 and 6. These were then pre-processed to give each student's results in terms of the 6 test item types; 5,5,4,4,3,3 were the number of items of these type 1 to 6 respectively. The particular form of the pre-processing depends on the item type values used: with the 0-N type node values, a student's results might be 541233, whereas with the H/M/L values, the same students results would be represented as HHLMHH.

The expert rule classifications were used to generate the priors for the sub-classifications. All the CPTs of the test item types take the form of $P(Type = Value|Classification = X)$. As we have seen from the domain description, the experts expect particular classes of students to get certain item types correct, and others wrong. However we do need to model the natural deviations from such "rules", where students make a careless error, that is, they apply their own logic but do not carry it through. For example, students who are thinking according to the LWH misconception are predicted to get all 5 items of Type 2 correct. If however that there is a probability of 0.1 of a careless mistake on any one item, the probability of a score of 5 is $(0.9)^5$, and the probability of other scores follows the binomial distribution; the full vector for P(Type2|Subclass=LWH) is (0.59,0.33,0.07,0.01,0.00,0.00) (to two decimal places). When the item type values H/M/L are used, the numbers are accumulated to give the vector (0.92,0.08,0.00) for H, M and L. We note that the experts consider that this mistake probability is considerably less than 0.1, say of the order of 1-2%.

Much more difficult than handling the careless errors in the well understood behaviour of the specific known misconceptions, is to model situations where the experts do not know how a student will behave. This was the case where the experts specified '.' for the classifications LU, SU, AU and UN in Table 1. We modelled the expert not knowing what such a student would do on the particular item type in the BN by using 0.5 (i.e. 50/50 that a student will get each item correct) with the binomial distribution to produce the CPTs.

We ran experiments with different probabilities for a single careless mistake ($pcm$=0.03, 0.11 and 0.22), with the CPTs calculated in this manner, to investigate the effect of this parameter on the behaviour of the system. These number were chosen to give a combined probability for HIGH (for 5 items) of 0.99, 0.9 and 0.7 respectively, numbers that our experts thought were reasonable. Results are described in Section 3.5.

### 3.4 BN EVALUATION PROCESS

During the expert elicitation process we performed the following three basic types of evaluation. First was

---

[2] An indication as to the meaning of these additional nodes is as follows. The "HN" nodes relate to the Hidden Numbers game, with evidence entered for the number of doors opened before an answer was given, and a measure of the "goodness of order" in opening doors. The root node for the Hidden Number game subnet reflects a player's game ability – in this case door opening "efficiency".

*Case-based evaluation*, where the experts "play" with the net, imitating the response of a student with certain misconceptions and review the posterior distributions on the net. Depending on the BN parameters, it was often the case that while the incorporation of the evidence for the 6 item types from the DCT test data greatly increased the BN's belief for a particular misconception, the expert classification was not the BN classification with the highest posterior, because it started with a low prior. We found that it was useful to the experts if we also provided the ratio by which each classification belief had changed (although the highest posterior is used in all evaluations).

During the use of the BN in the full ITS (see Figure 1), each time student answers are entered, the posteriors for the fine classification are updated and in turn become the new priors for that node; in this way, the network adapts to the individual student over a range of games and item types over time. This adaptive aspect allows the system to identify students with misconceptions that are fairly infrequent in the overall population. This motivated our *Adaptiveness evaluation*, where the experts imitate repeated responses of a student, update the priors after every test and enter another expected test result. This detection of classifications over repetitive testing built up the confidence of the experts in the adaptive use of the BN.

Next, we undertook *Comparison evaluation* between the classifications of the BN compared to the expert rules on the DCT data.[3] As well as a comparison grid (see next subsection), we provided the experts with details of the records where the BN classification differed from that of the expert rules. This output proved to be very useful for the expert in order to understand the way the net was working and to build their confidence in the net.

Finally, we undertook a *Prediction evaluation* which considers the prediction of student performance on individual item type nodes rather than direct misconception diagnosis. We enter a student's answers for 5 of the 6 item type nodes, then predict their answer for the remaining one; this is repeated for each item type. The number of correct predictions gives a measure of the accuracy of each model, using a score of 1 for a correct prediction (using the highest posterior) and 0 for an incorrect prediction. We also look at the predicted probability for the actual student answer. Both measures are averaged over all students.[4]

We performed these four types of evaluation every time we modified the BN structure or the CPTs. This way we were aware of the implications this change may have had on the overall behaviour. The iterative process halts when the experts feel the behaviour of the BN is satisfactory.

---

[3] We note that this evaluation is similar to the comparison used in [van der Gaag et al., 2000].

[4] This evaluation method was suggested by an anonymous reviewer; the analysis of results using these prediction measures is preliminary due to time constraints.

Table 2: Expert rule vs expert elicited BN classification. Type node states 0-N, $pcm=0.11$.

|     | lwh | lze | lrv | lu | sdf | srn | su | ate | amo | mis | au | un |
|-----|-----|-----|-----|----|-----|-----|----|----|----|----|----|----|
| lwh | 386 | 0 | 0 | 0 | 0 | 0 | 0 | 0 | 0 | 0 | 0 | 0 |
| lze | 0 | 98 | 0 | 0 | 0 | 0 | 0 | 0 | 0 | 0 | 0 | 0 |
| lrv | 10 | 0 | 0 | 0 | 0 | 0 | 0 | 0 | 0 | 0 | 0 | 0 |
| lu | 6 | 9 | 0 | 54 | 0 | 0 | 0 | 0 | 0 | 0 | 0 | **6** |
| sdf | 0 | 0 | 0 | 0 | 83 | 0 | **4** | 0 | 0 | 0 | 0 | 0 |
| srn | 0 | 0 | 0 | 0 | 0 | 159 | 0 | 0 | 0 | 0 | 0 | 0 |
| su | 0 | 0 | 0 | 0 | *2* | *22* | 40 | **3** | 0 | 0 | 0 | **2** |
| ate | 0 | 0 | 0 | 0 | 0 | 0 | 0 | 1050 | 0 | 0 | 0 | 0 |
| amo | 0 | 0 | 0 | 0 | 0 | 0 | 0 | 0 | 79 | 0 | 0 | 0 |
| mis | 0 | 0 | 0 | 0 | 0 | 0 | 0 | 0 | 0 | 6 | 0 | 0 |
| au | **9** | 0 | 0 | 0 | 0 | 0 | 0 | *63* | *8* | 0 | 0 | **1** |
| un | *43* | *6* | 0 | *15* | *35* | *14* | *11* | *119* | *26* | *2* | 0 | 66 |

### 3.5 RESULTS

Table 2 is an example of the comparison grids for the fine classification that were produced during the comparison evaluation phase. Similar grids were produced for the coarse classification. Each row corresponds to the expert rules classification, while each column corresponds to the BN classification, using the highest posterior; each entry in the grid shows how many students had a particular combination of classifications from the two methods. The grid diagonals show those students for whom the two classifications are in agreement, while the "desirable" changes are shown in *italics*, and undesirable changes are shown in **bold**. Note that we use the term "match", rather than saying that the BN classification was "correct", because the expert rule classification is not necessarily ideal.

Further assessment of these results by the experts revealed that when the BN classification does not match the expert rules classification, the misconception with the second highest posterior often did match. The experts then assessed whether differences in the BN's classification from the expert rules classification were in some way desirable or undesirable, depending on how the BN classification would be used. They came up with the following general principles which provided some general comparison measures: (1) it is *desirable* for expert rule classified LUs to be re-classified as another of the specific Ls, similarly for AUs and SUs, and it was desirable for Us to be re-classified as anything else; (because this is dealing with borderline cases that the expert rule really can't say much about); (2) it is *undesirable* for (a) specific classifications (i.e. not those involving any kind of "U") to change, because the experts are confident about these classifications, and (b) for any classification to change to UN, because this is in some sense throwing away information (e.g. LU to UN loses information about the "L-like" behaviour of the students).

Table 3: Coarse classification grid, expert rules vs expert elicited BN, varying $pcm$ (0.22, 0.11, 0.03), and item type values (H/M/L and 0-N).

| | 0-N | | | | H/M/L | | | |
|---|---|---|---|---|---|---|---|---|
| | A | S | L | UN | A | S | L | UN |
| 0.22 | 86.95% | 12.56% | 0.49% | | 87.61% | 11.98% | 0.41% | |
| A | 1207 | 0 | 9 | 0 | 1213 | 0 | 0 | 3 |
| S | 3 | 312 | 0 | 0 | 4 | 310 | 0 | 1 |
| L | 0 | 0 | 569 | 0 | 0 | 0 | 557 | 2 |
| UN | 157 | 71 | 78 | 31 | 150 | 82 | 60 | 45 |
| 0.11 | 88.02% | 11.12% | 0.86% | | 87.28% | 10.71% | 2.01% | |
| A | 1206 | 0 | 9 | 1 | 1184 | 0 | 23 | 9 |
| S | 3 | 310 | 0 | 2 | 4 | 310 | 0 | 1 |
| L | 0 | 0 | 563 | 6 | 7 | 0 | 557 | 5 |
| UN | 147 | 60 | 64 | 66 | 139 | 73 | 49 | 76 |
| 0.03 | 89.29% | 9.48% | 1.23% | | 91.63% | 5.25% | 3.12% | |
| A | 1202 | 0 | 8 | 6 | 1173 | 0 | 0 | 43 |
| S | 3 | 308 | 0 | 4 | 0 | 304 | 0 | 11 |
| L | 0 | 0 | 560 | 9 | 0 | 0 | 547 | 22 |
| UN | 102 | 49 | 80 | 106 | 83 | 9 | 36 | 209 |

Table 4: Fine classification summary comparison various models compared to the expert rules.

| Method | Type values | | Match | Des. change | Undes. change |
|---|---|---|---|---|---|
| Expert BN | 0-N | 0.22 | 77.88 | 20.39 | 1.72 |
| | | 0.11 | 82.93 | 15.63 | 1.44 |
| | | 0.03 | 84.37 | 11.86 | 3.78 |
| | H/M/L | 0.22 | 80.47 | 18.71 | 0.82 |
| | | 0.11 | 83.91 | 13.66 | 2.42 |
| | | 0.03 | 90.40 | 6.48 | 3.12 |
| SNOB | 24 DCT | | 79.81 | 17.60 | 2.49 |
| | 0-N | | 72.06 | 16.00 | 11.94 |
| | H/M/L | | 72.51 | 17.03 | 10.46 |
| EBN learned | 0-N | Avg | 95.97 | 2.36 | 1.66 |
| | H/M/L | Avg | 97.63 | 1.61 | 0.75 |
| CaMML constr. | 0-N | Avg | 86.51 | 5.08 | 8.41 |
| | H/M/L | Avg | 83.48 | 8.12 | 8.34 |
| CaMML uncons. | 0-N | Avg | 86.15 | 5.87 | 7.92 |
| | H/M/L | Avg | 92.63 | 4.61 | 2.76 |

Table 3 shows the coarse classification comparison grids obtained when varying the probability of a careless mistake ($pcm$=0.22, 0.11 and 0.03) and the item type values (0-N vs H/M/L). Each grid is accompanied by the percentages for match, desirable and undesirable change. As the probability decreases, the total number of matches with expert classifications goes up, due to more UN students being in agreement, however more L,S and A students no longer match, which is considered "undesirable" (see above). In effect, the definition of A,S, and L becomes more stringent as the probability of a careless error decreases, so more move out of A, L and S into UN, and less move from UN into A, L and S. There are also shifts between undesirable classification differences, for example the 8 A students who the BN classifies as L (values 0-N, $pcm$=0.22) (in fact, highly offensive to our experts!), shift to the also generally undesirable UN ($pcm$=0.03).

We believe that the differences between the BN and the expert rule classifications are due to the following factors. First, the expert rules give priority to the type 1 and type 2 results, whereas the BN model gives equal weighting to all 6 item types. An example of this is a student with answers 450433, who the expert rule classifies as AU due to the "High" result for item types 1 and 2 (ignoring the other answers). The BN with a fairly high chance of a careless mistake (0.22) says this student looks like an LHW (050433), as the only difference is the answer for item type 1, while for $pcm$=0.03, the BN classifies the student as UN. Second, the expert elicited BN structure and parameters reflects both the experts' good understanding for the known fine classifications, and their poor understanding of the behaviour of "U" students (LU, SU, AU, and UN). Finally, as discussed earlier, some classes are broken down into fine classifications more than others, resulting in lower priors, so the more common classifications (such as ATE and UN) tend to draw in others.

Closer inspection also shows that some "undesirable" changes are reasonable. For example a student answering 443322 is classified as an ATE by both the expert rule and the H/M/L BN, since one mistake on any item is considered "high". However the 0-N BN (for $pcm$=0.03) classifies the student as UN, since the combined probability of 5 careless mistakes (one on each item type) is very low.

It is not possible for reasons of space to present the full set of results for the fine classifications. Table 4 (Set 1) shows a summary of the expert BN fine classification, varying the type values and probability of careless mistake, in terms of percentage of matches (i.e. on the grid diagonal), desirable changes and undesirable changes. We can see that matches are higher for H/M/L than the corresponding O-N run, desirable changes are lower, while there is no consistent trend for undesirable changes. The undesirable change percentages are quite low, especially considering that we know some of these can be considered quite justified. Table 5 (Set 1) shows the two prediction measures (see Section 3.4), averaged over all predicted item types, for all students. Both measures show the H/M/L model giving better prediction results than the corresponding 0-N run. Both measures show the probability of a careless error effects the results for the 0-N models, but only the predicted probability shows an effect on the H/M/L model results.

Table 5: Accuracy of various models predicting student item type answers.

| Method | Type values | | Avg Pred. Accuracy | Avg Pred. Prob. |
|---|---|---|---|---|
| Expert BN | 0-N | 0.22 | 0.34 | 0.34 |
| | | 0.11 | 0.83 | 0.53 |
| | | 0.03 | 0.82 | 0.70 |
| | H/M/L | 0.22 | 0.89 | 0.69 |
| | | 0.11 | 0.89 | 0.80 |
| | | 0.03 | 0.88 | 0.83 |
| EBN learned | 0-N | Avg | 0.83 | 0.74 |
| | H/M/L | Avg | 0.89 | 0.83 |
| CaMML constr. | 0-N | Avg | 0.83 | 0.72 |
| | H/M/L | Avg | 0.88 | 0.79 |
| CaMML uncons. | 0-N | Avg | 0.83 | 0.74 |
| | H/M/L | Avg | 0.89 | 0.83 |

Overall it is clear that the expert elicited network performs a good classification of students misconceptions, and captures well the different uncertainties in the experts domain knowledge. In addition, its performance is quite robust to changes in parameters such as the probability of careless mistakes or the granularity of the evidence nodes.

## 4 KNOWLEDGE DISCOVERY

The next stage of the project involved the application of certain automated methods for knowledge discovery to the domain data.

### 4.1 CLASSIFICATION

The first aspect investigated was the classification of decimal misconceptions. We applied the SNOB classification program[Wallace and Dowe, 2000], based on the information theoretic Minimum Message Length (MML). SNOB was run on the data from 2437 students on 24 DCT items, each being a binary value as to whether the student got the item correct or incorrect, with a variety of initial guesses for the number of classes (5,10,15,20,30). All five classifications were very similar; we present here results from the model with the lowest MML estimate (5 initial classes). Using the most probable class for each member, we constructed a grid comparing the SNOB classification with the expert rule classification. Of the 12 classes produced by SNOB, we were able to identify 8 that corresponded closely to the expert classifications (i.e. had most members on the grid diagonal). Two classes were not found (LRV and SU). Of the other 4 classes, 2 were mainly combinations of the AU and UN classifications, while the other 2 were mainly UNs. SNOB was unable to classify 15 students (0.6%). The percentages of match, desirable and undesirable change are shown in Table 4 (set 2, row 1). They are comparable with the expert BN 0-N and only slightly worse than the expert BN H/M/L results.

SNOB was then run on the pre-processed data consisting of student answers on the 6 item types (values 0-N and H/M/L). The comparison results for this run were not particularly good. For O-N type values, SNOB found only 5 classes (32 students = 1.3% not classified), corresponding roughly to some of the most populous expert classes – LWH, SDF, SRN, ATE and UN, and subsuming the other expert classes. For H/M/L type values, SNOB found 6 classes (33 students = 1.4% not classified), corresponding roughly to 5 of the most populous expert classes (LWH, SDF, SRN, ATE, UN), plus a class that combined MIS with UN. In this case LZEs were all grouped with ATEs, as were AMOs. The match results are shown in Table 4 (set 2, rows 2 and 3). Clearly, summarising the results of 24 DCT into types gives relatively poor performance; it is proposed that this is because many pairs of the classes are distinguished by student behaviour on just one item type, and SNOB might consider these differences to be noise within one class.

The overall good performance of the classification method shows that automated knowledge discovery methods may be useful in assisting expert identify suitable values for classification type variables.

### 4.2 PARAMETERS

Our next investigation was to learn the parameters for the expert elicited network structure. The data was randomly divided into five 80%-20% splits for training and testing; the training data was used to parameterise the expert BN structures using the Netica BN software's parameter learning feature[5], while the test data was given to the resultant BN for classification. The match results (averaged over the 5 splits) for the fine classification comparison of the expert BN structures (with the different type values, 0-N and H/M/L) with learned parameters are shown in Table 4 (set 3), with corresponding prediction results (also averaged over the 5 splits) given shown in Table 5 (set 2).

The average prediction probabilities for the BN with learned parameters are better than for the expert BNs for the O-N type values (0.74 compared to 0.70); the other prediction results show no significant difference. The match percentages for both BNs with learned parameters are significantly higher than for all the expert BNs with elicited parameters (with varying $pcm$), while both the desirable and undesirable changes are lower; most of the difference is due to the reduction in desirable changes. Within the learned parameter results, the match percentage is significantly higher for H/M/L than 0-N, while the changes are lower. In both cases, the percentage of undesirable changes is lower than the desirable change. Clearly, learning the pa-rameters from data, if it is available, gives results that are much closer to the expert rule classification. The trade-off is that the network no longer makes changes to the various "U" classifications, i.e. it doesn't shift LUs, SUs, AUs and UNs into other classifications that may be more useful in a teaching context. However it does mean that expert input into the knowledge engineering process can be reduced, with the parameter learning on an elicited structure giving a BN model that can be used in the ITS.

---
[5]www.norsys.com

### 4.3 STRUCTURE

Our third investigation involved the application of Causal MML (CaMML) [Wallace and Korb, 1999] to learn network structure. In order to compare the learned structure with that of the expert elicited BN, we decided to use the pre-processed 6 type data; each program was given the student data for 7 variables (the fine classification variable and the 6 item types), with both the 0-N values and the H/M/L values. The same 5 random 80%-20% splits of the data were used for training and testing. The training data was given as input to the structural learning algorithm, and then used to parameterise the result networks using Netica's parameter learning method.

We ran CaMML once for each split (a) without any constraints and (b) with the ordering constraint that the classification should be an ancestor of each of the type nodes. This constraint reflects the general known causal structure. Each run produced a slightly different network structure, with some having the fineClass node as a root, some not. One fairly typical network with the ordering constraint contained 4 arcs from the class node to type nodes, with one type node also being a root node, only two type nodes leaves, and 10 arcs between type nodes. The arcs/nodes ratio of the learned structures varies from 1.4 to 2.2, while the number of parameters varies from about 700 to 144,000; the structures produced for the H/M/L data seem simpler using these measures, but this is not statistically significant.

The percentage match results comparing the CaMML BN classifications (constrained and unconstrained, 0-N and H/M/L) are also shown in Table 4 (sets 4 and 5), with the prediction results shown in Table 5 (sets 3 and 4). The prediction results for both 0-N and H/M/L are similar to those of the fully elicited expert BNs. The match percentages are similar to those of the fully elicited expert BN for the 0-N, however the undesirable change percentages are higher, while the desirable change percentages are lower. For H/M/L, the match results are higher than for the expert BN (92.63 compared to the highest of 90.40), with fewer desirable and undesirable changes.

The undesirable changes include quite a few shifts from one specific classification to another, which is particularly bad as far as our experts are concerned; for example, several of the networks do not identify the SDF and MIS classifications, instead grouping them with ATE. We also note that the variation between the results for each data set 1-5 was much higher than for the variation when learning parameters for the expert BN structure. This no doubt reflects the difference between the network structure learned for the different splits. However we did not find a clear correlation between the complexity of the learned network structures and their classification performance.

In seeking to improve automated discovery of structure by exploiting expert domain knowledge, experts could provide constraints to guide the search and could manually select for further investigation those alternative structures which were best interpretable in terms of the domain concepts.

## 5  CONCLUSIONS

This work began with the recognition that we had access to a novel combination of data and information which could enable the developments and comparative studies reported above: a domain where student misconceptions abound; involvement of experts with a detailed understanding of the basis of the misconceptions, and how they relate to domain specific activities; and an extensive data set of student behaviour on test items in the domain. The work reported here falls into three components: (a) the development, by expert elicitation, of a Bayesian network designed to be the "engine" of an adaptive tutoring system; an elicitation strongly informed by the experts' detailed understanding of patterns in the data set; (b) a study of learning techniques applied to the same data set – looking at learning of classification, structure, and parameters – which could be compared against the experts' network; and (c) some reflection, based on the experience of working with the experts and the automated tools, as to how elicitation and knowledge discovery might be combined in the BN knowledge engineering process.

A first, important, observation is that the automated techniques were able to yield networks which gave quantitative results comparable to the results from the BN elicited from the experts. This level of matching provides a form of 'validation' of the learning techniques and suggests that automated methods can reduce the input required from domain experts. It also supports the reciprocal conclusion regarding the validity of manual construction when there is enough expert knowledge but no available data set. In addition, we have seen that the use of automated techniques can provide opportunities to explore the implications of modelling choices and to get a feel for design tradeoffs – some examples of this were reported above in both the initial elicitation stage, and the discovery stage (e.g. 0-N *vs.* H/M/L).

Given that elicited BN was based on the expert knowledge that had been accumulated over a period of time through much analysis and investigation, how useful is an automated approach in domains where such detailed (validation) knowledge is not available? Our experience suggests that a hybrid of expert and automated approaches is feasible. We plan to apply these methods in a situation (student work on algebra) where we have data on student behaviour, but do not have detailed prior expert analysis of the data.

**Acknowledgements** The authors would like to acknowledge the preliminary work undertaken by Elise Dettman, and thank Brent Boerlage for his assistance with Netica and the anonymous reviewers for their helpful suggestions.